\documentclass[11pt,a4paper]{article}
\usepackage[hyperref]{eacl2021}
\usepackage{times}
\usepackage{latexsym}
\usepackage{multirow}
\usepackage{array}
\usepackage{graphicx}
\usepackage{changes}

\usepackage{microtype}

\aclfinalcopy % Uncomment this line for the final submission
\makeatletter
\def\blfootnote{\xdef\@thefnmark{}\@footnotetext}
\makeatother

\title{AnswerQuest: A System for Generating Question-Answer Items from Multi-Paragraph Documents}

\author{Melissa Roemmele$^\dagger$ \quad Deep Sidhpura$^\Diamond$ \quad Steve DeNeefe$^\dagger$ \quad Ling Tsou$^\dagger$\\
SDL Research, Los Angeles, CA, USA\\
$^\dagger$\texttt{\string{mroemmele,sdeneefe,ltsou\string}@sdl.com}\\
$^\Diamond$\texttt{deepsidhpura777@gmail.com}}

\date{}

\begin{document}
\maketitle
\begin{abstract}
\blfootnote{$^\Diamond$Current affiliation: eBay Inc., San Jose, CA, USA}
One strategy for facilitating reading comprehension is to present information in a question-and-answer format. We demo a system that integrates the tasks of question answering (QA) and question generation (QG) in order to produce Q\&A items that convey the content of multi-paragraph documents. We report some experiments for QA and QG that yield improvements on both tasks, and assess how they interact to produce a list of Q\&A items for a text. The demo is accessible at \href{https://qna.sdl.com/}{qna.sdl.com}.
\end{abstract}

\section{Introduction}

%\deleted{}
%\replaced{}{}
%\added{{}

Automated reading comprehension is one of the current frontiers in AI and NLP research, evidenced by the frequently changing state-of-the-art among competing approaches on standard benchmark tasks \cite[e.g.][]{wang-etal-2018-glue}. These systems aim to reach the standard of human performance, but they also have the potential to further enhance human reading comprehension. For instance, many demonstrations of reading comprehension involve eliciting answers to questions about a text. Meanwhile, educational research and conventional writing advice indicate that structuring information in a question-and-answer format can aid comprehension \cite{knight2010,raphael1982improving}. Accordingly, systems that present content in this format by automatically generating and answering relevant questions may help users better understand the content.

The two NLP tasks essential to this objective, question answering (QA) and question generation (QG), have received a lot of attention in recent years. Recent work has started to explore the intersection of QA and QG for the purpose of enhancing performance on one or both tasks \cite{sachan-xing-2018-self,song-etal-2018-leveraging,tang-etal-2018-learning,yuan-etal-2017-machine}. Among application interfaces that demo these tasks, most have focused on either one or the other \cite{kaisser-2008-qualim,kumar2019paraqg}. \citet{squash2019} presented a system that integrated these tasks to simulate a pedagogical approach to human reading comprehension. In our work, we demo an end-to-end system that applies QA and QG to multi-paragraph documents for the purpose of user content understanding. The system generates a catalog of Q\&A items that convey a document's content. This paper first presents some focused contributions to the individual tasks of QA and QG. In particular, we examine the challenging task of QA applied to multi-paragraph documents and show the impact of incorporating a pre-trained text encoding model into an existing approach. Additionally, we report a new set of results for QA that assesses generalizability between datasets that are typically evaluated separately. For QG, we demonstrate the benefit of data augmentation by seeding a model with automatically generated questions, which produces more fluent and answerable questions beyond a model that observes only the original human-authored data. In combining the two tasks into a single pipeline, we show that the information given by the generated Q\&A items is relevant to the information humans target when formulating questions about a text. 

The demo is implemented as a web application in which users can automatically generate Q\&A pairs for any text they provide. The web application is available at \href{https://qna.sdl.com/}{qna.sdl.com}, and our code is at \href{https://github.com/roemmele/answerquest}{github.com/roemmele/answerquest}.

%, and also obtain answers for their own custom questions about the text. 

% We describe the design, experiments, and evaluations we conducted for the QA (Section \ref{qa}) and QG (Section \ref{qg}) tasks, as well as their integration in order to produce Q\&A pairs (Section \ref{qna_section}). 
% 
% \deleted{Within each section we highlight the relevant existing work motivating our exploration. Then we describe the behavior and interaction of these components within the demo interface (Section \ref{demo_interaction}) prior to some concluding comments about future objectives (Section \ref{conclusion}).} 

\section{Question Answering}\label{qa}

\subsection{Model Overview}\label{qa_overview}

Our demo implements extractive QA, where answers to questions are extracted directly from some given reference text. State-of-the-art systems have utilized a classification objective to predict indices of answer spans in the text. This approach has achieved success when the reference text is limited to a single paragraph \cite{devlin-etal-2019-bert}.  However, QA for multi-paragraph documents has proven to be more difficult. Our system addresses this challenging document-level QA task by adapting an existing method to additionally leverage a pre-trained text encoding model. 

Existing work on document-level QA has proposed a pipelined approach that first applies a retrieval module to select the most relevant paragraphs from the document, and then a reader module for extracting answers from the retrieved paragraphs \cite{chen-etal-2017-reading,yang-etal-2019-end-end-open}. During training, each of the retrieved paragraphs and the corresponding questions are observed independently. To predict answers, the model scores candidate answer spans within each of the paragraphs, ultimately predicting the one with the highest score across all paragraphs. One problem is that the candidate answer scores across paragraphs might not be comparable, since each paragraph-question pair is treated as an independent training instance. To address this issue, \citet{clark-gardner-2018-simple} suggested a shared-normalization approach (which we refer to here as \textsc{BiDAF Shared-Norm}) where paragraph-question pairs are still processed independently, but answer probability scores are globally normalized across the document. In their work, they selected the top-k most relevant paragraphs for a given question using a TF-IDF heuristic. They then encoded the question and these paragraphs into a neural architecture consisting of GRU layers and a Bi-Directional Attention Flow (BiDAF) mechanism \cite{seo2017}. On top of this model is a linear layer that predicts the start and end token indices of the answer within a paragraph, using an adapted softmax function with normalization across all top-k paragraphs for the question. 

Another document-level QA system, \textsc{RE$^3$QA} \cite{hu-etal-2019-retrieve}, incorporated the text encoding model \textsc{BERT} \cite{devlin-etal-2019-bert}. \textsc{BERT} has been successfully used for numerous reading comprehension tasks. In contrast to \textsc{BiDAF Shared-Norm}, \textsc{RE$^3$QA} combined paragraph retrieval and answer prediction into a single end-to-end training process, applying \textsc{BERT} to both steps. Because it obtained favorable results relative to the \textsc{BiDAF Shared-Norm} approach, we were curious to assess the isolated impact of \textsc{BERT} specifically on the answer prediction component of the pipeline. Therefore we adapted \citeauthor{clark-gardner-2018-simple}'s shared-normalization approach by replacing their GRU BiDAF encoder with the \textsc{BERT-Base-Uncased} encoder. \citet{wang2019multi} used a similar approach for open-domain QA, where answers are mined from the entirety of Wikipedia. We instead evaluate QA with reference to a single document, for which the impact of BERT on the shared-normalization approach has not yet been documented.

We refer to our model here as \textsc{BERT Shared-Norm}. To rank paragraph relevance to a question, we rely on TF-IDF similarity. During training, we retrieved the top k=4 paragraphs. The \textsc{BERT Shared-Norm} model consists of the \textsc{BERT-Base-Uncased} pre-trained model, which encodes the paragraph and question in the same manner as \citeauthor{devlin-etal-2019-bert}'s paragraph-level QA model. The rest of our model is the same as \textsc{BiDAF Shared-Norm}: the softmax output layer predicts the start and end answer tokens and the same shared-normalization objective function is applied during training. The model can predict that a question is `unanswerable' by observing an index of 0 for the end token. During inference, the highest-scoring answer span across paragraphs is predicted as the answer. See Appendix \ref{qa_details} for more details.

\subsection{Dataset}

Our QA experiments utilized the \textsc{SQuAD} \cite{rajpurkar-etal-2016-squad} and \textsc{NewsQA} \cite{trischler-etal-2017-newsqa} datasets. \textsc{SQuAD} is derived from Wikipedia articles, while \textsc{NewsQA} consists of CNN news articles. Both datasets were developed through crowdsourcing tasks where participants authored questions and identified their answers, resulting in text-question-answer items where each answer is a span within the text. There are two versions of \textsc{SQuAD}. \textsc{SQuAD-1.1} contains 87,599 train and 10,570 test items. \textsc{SQuAD-2.0} contains an additional 42,720 train and 1,303 test items (a total of 130,319 and 11,873, respectively), distinguished from \textsc{SQuAD-1.1} by including questions that do not have answers in the text. \textsc{NewsQA} contains 107,674 and 5,988 train and test items, respectively. As with \textsc{SQuAD-2.0}, some of these questions are unanswerable.\footnote{The \textsc{SQuAD} test items we use are actually the items from their `dev' (development) set: \href{https://rajpurkar.github.io/SQuAD-explorer/}{rajpurkar.github.io/SQuAD-explorer}. Their official test set is withheld. The other published systems we compare against also report evaluations on this dev set, so for simplicity we refer to it here as the test set. Similarly, we use the dev \textsc{NewsQA} items as our held-out test set: \href{https://github.com/Maluuba/newsqa}{github.com/Maluuba/newsqa}.}

% As with SQuAD, the evaluation items are actually labeled as the `development' set that we here we utilize as a held-out test set.

\textsc{SQuAD} questions pertain to a single paragraph. Paragraphs are grouped by document and can be concatenated for document-level QA. There are on average 43 paragraphs per document. Paragraph boundaries are not explicit in the \textsc{NewsQA} texts, so we treated each text as a multi-paragraph document by splitting it into chunks of 300 tokens, resulting in 2.55 average paragraphs per document.

\subsection{Evaluation}

\subsubsection{Comparison with other Systems}\label{qa_comparison_section}

We first evaluated our \textsc{BERT Shared-Norm} model on \textsc{SQuAD-1.1} for comparison with the \textsc{BiDAF Shared-Norm} and \textsc{RE$^3$QA} results reported for this dataset. We used the official \textsc{SQuAD} evaluation scripts provided by the website. For direct comparison with \textsc{BiDAF Shared-Norm}, we replicated their setting of k=15 for paragraph retrieval. Table \ref{qa_comp_eval} shows the results in terms of the exact match (EM) and F1 accuracy of answers. We improve upon the result for \textsc{BiDAF Shared-Norm}, demonstrating the beneficial impact of incorporating \textsc{BERT} into this approach. The \textsc{BERT}-based \textsc{RE$^3$QA} still outperforms our model, suggesting that its other components outside the \textsc{BERT} encoding for answer prediction additionally contribute to its success.

\begin{table}[th!]
\begin{center}
\begin{tabular}{ | l | l | l | }
\hline
\textbf{Model} & \textbf{EM} & \textbf{F1} \\
\hline
\textsc{BiDAF Shared-Norm} & 64.08 & 72.37 \\
\textsc{RE$^3$QA} & 77.90 & 84.81 \\
\textsc{BERT Shared-Norm} & 72.85 & 80.58 \\
\hline
\end{tabular}
\caption{QA results on SQuAD-1.1}
\label{qa_comp_eval}
\end{center}
\end{table}

\subsubsection{Generalizability across Datasets}

%Moreover, we want our system to recognize when questions are not unanswerable, which may be particularly challenging given that models may over- or under-predict unanswerability when observing text from new domains. 

Our demo accepts any arbitrary text supplied by a user, and we ultimately aim to produce informative Q\&A items for varying content domains. State-of-the-art QA systems have matched human-level performance on individual datasets like \textsc{SQuAD}, but it is unclear how much this performance generalizes across different datasets. As a narrow assessment of this issue, we examined the generalizability between \textsc{SQuAD} and \textsc{NewsQA} by alternatively training and evaluating \textsc{BERT Shared-Norm} on different combinations of these datasets.

Table \ref{qa_gen_eval} shows the results of this experiment. We trained three different \textsc{BERT Shared-Norm} models on separate datasets: \textsc{SQuAD-2.0}, \textsc{NewsQA}, and \textsc{SQuAD-2.0} + \textsc{NewsQA} combined (which we term \textsc{MegaQA}). We then evaluated each of these models on the \textsc{SQuAD-2.0} and \textsc{NewsQA} test sets. Note that the experiments in Section \ref{qa_comparison_section} were evaluated on \textsc{SQuAD-1.1} for comparison with the other approaches. Here, we only evaluate on \textsc{SQuAD-2.0}, which involves additionally predicting when a question does not have an answer span in the document. For these evaluations, consistent with training, we retrieved the top k=4 paragraphs from each document for answer prediction.

\begin{table}[th!]
\begin{center}
\begin{tabular}{ | l | l | l | l | l |}
\hline
\multirow{3}{*}{\textbf{Train Data}} & \multicolumn{4}{c|}{\textbf{Test Data}}\\
\cline{2-5}
& \multicolumn{2}{c|}{\textbf{\textsc{SQuAD-2.0}}} & \multicolumn{2}{c|}{\textbf{\textsc{NewsQA}}}\\
\cline{2-5}
& \textbf{EM} & \textbf{F1} & \textbf{EM} &\textbf{F1} \\
\hline
\textsc{SQuAD-2.0} & 71.37 & 74.65 & 40.88 & 48.67\\
\textsc{NewsQA} & 45.85 & 49.88 & 52.68 & 61.26\\
\textsc{MegaQA} & 70.29 & 73.55 & 53.85 & 62.46\\
\hline
\end{tabular}
\caption{Generalizability of \textsc{BERT Shared-Norm} across datasets}
\label{qa_gen_eval}
\end{center}
\end{table}

The results reveal a generalizability problem, where the model trained on \textsc{SQuAD-2.0} fails to perform as well on \textsc{NewsQA} and vice-versa, presumably due to their domain difference (Wikipedia vs. Newswire). However, combining the datasets with the \textsc{MegaQA} model generalizes well to both. Related to this, \citet{talmor-berant-2019-multiqa} found combining multiple datasets from different domains to be advantageous for \textsc{BERT}-based reading comprehension models. Based on these results, the \textsc{BERT Shared-Norm MegaQA} model is currently integrated into our demo.

\section{Question Generation}\label{qg}

\subsection{Model Overview}\label{qg_model_overview}

We follow the same paradigm of much recent work on QG, which has applied encoder-decoder (i.e. sequence-to-sequence) models to text-question pairs \cite{du-etal-2017-learning,duan-etal-2017-question,scialom2019self,song-etal-2018-leveraging,zhao-etal-2018-paragraph}. Similar to \citeauthor{scialom2019self}, we utilize the Transformer architecture for the encoder and decoder layers of the model, and enhance the decoder with a copy mechanism.
% \deleted{ with the specific variation of using the Transformer \cite{vaswani2017attention} architecture for the encoder and decoder layers, instead of RNN layers as used in the previous work.}
The encoder input is a single sentence and the decoder output is a question, where the input sentence contains the answer to the question. Following the standard procedure for sequence-to-sequence model training, we used the cross-entropy of the output question tokens as the loss function. When generating questions, we use a beam size of 5. See Appendix \ref{qg_details} for further details.

\subsection{Dataset}\label{qg_dataset}

We trained and evaluated the model on  \textsc{SQuAD} and \textsc{NewsQA} concatenated, the same datasets used for the QA experiments. Our QG model aims to produce questions whose answers are contained in their corresponding input texts, so we only included \textsc{SQuAD-1.1} items and answerable \textsc{NewsQA} items (this excluded 32,764 \textsc{NewsQA} items from the train and test sets). For each paragraph-question-answer item, we sentence-segmented the paragraph, isolated the sentence with the answer span, and inserted special tokens into the sentence (\texttt{$<$ANSWER$>$} and \texttt{$<$/ANSWER$>$}) designating the start and end of the span. These answer-annotated sentences were the model inputs and the aligned questions were the target outputs. We applied Byte-Pair-Encoding (BPE) tokenization \cite{sennrich2016neural} to the inputs and targets (see Appendix \ref{qg_details}). We used the same train-test dataset splits as the QA experiments, allocating a small subset of training items to a validation set for hyperparameter tuning. Overall the train, validation, and test sets consisted of 160,876, 3,281, and 14,910 sentence-question pairs, respectively.

\subsection{Data Augmentation Experiments}

We examined three different versions of the model described in \ref{qg_model_overview}, differentiated by their training inputs. The purpose of this experiment was to assess using the output of a rule-based QG system as a means of augmenting the training data. We specifically evaluated the three configurations below:

\textbf{\textsc{Standard}}: In this model, no data augmentation was applied. We trained the model directly on the \textsc{SQuAD}/\textsc{NewsQA} items described in \ref{qg_dataset}.

\textbf{\textsc{RuleMimic}}: This model observed only the automatically generated augmentation data, without the original data. The source of the augmentation data was the QG system by \citet{heilman-smith-2010-good}\footnote{Code at \href{http://www.cs.cmu.edu/~ark/mheilman/questions}{cs.cmu.edu/$\sim$ark/mheilman/questions}}. This system applies linguistic rule-based transformations (i.e. clause simplification, verb decomposition, subject-auxiliary inversion, and wh-movement) to convert a sentence into a question answered by the sentence, then scores the fluency of the question using a statistical model. \citet{du-etal-2017-learning} found favorable results for a neural sequence-to-sequence approach relative to this rule-based system, but we were curious about its use as a strategy for augmenting our training data. We anticipated that a neural model could learn to `mimic' the system's generic transformation rules by observing its inputs and outputs. Thus, we applied the system to the raw paragraphs in the \textsc{SQuAD}/\textsc{NewsQA} training set, which resulted in 1,531,233 questions, each aligned with a sentence. We then followed the same steps described in \ref{qg_dataset} to tokenize the sentence and mark the answer span. The training set for this model consisted only of these automatically generated questions (1,500,610 train items with 30,623 used for validation), with no human-authored questions.

\textbf{\textsc{Augmented}}: This model observed both the original data seen by the \textsc{Standard} model and the augmentation data seen by the \textsc{RuleMimic} model, via a two-stage fine-tuning process. After training the \textsc{RuleMimic} model, we used its parameters to initialize another model, then fine-tuned this new model on the \textsc{Standard} model dataset. The hypothesis behind this approach is that it can simulate linguistic rules underlying question formulation, while also capturing the more abstractive features of human questions that are harder to derive using deterministic syntactic and lexical transformations.

\subsection{Evaluation}

Many QG systems are evaluated using BLEU or similar metrics that reward overall token overlap between generated and human-authored questions. However, \citet{nema2018towards} argue that these metrics are ill-suited for QG. In particular, comparatively fluent questions with the same answer could have few tokens in common. Moreover, certain tokens within a question have far more impact than others on its perceived quality. They encourage alternative metrics that focus instead on the `answerability' of questions. Guided by this, we conducted both automated and human ratings-based evaluations in order to assess the answerability of our QG output. Because our demo performs extractive QA, our evaluations focus on whether questions are answerable relative to the input text from which the question is generated.

\subsubsection{Automated Evaluation}

% \deleted{\citet{duan-etal-2017-question} used a QA system as a training reward function for QG in order to promote generated questions that yielded the same answers they were conditioned upon.} 
Some work has utilized automated QA as a scoring metric for QG systems, based on the rationale that a QA system's ability to predict correct answers to generated questions indicate how well the questions are formulated to elicit these answers \cite{duan-etal-2017-question,zhang2019addressing}. Following this idea, we generated questions for sentence inputs in the \textsc{SQuAD}/\textsc{NewsQA} test set. As with the training inputs, these inputs were derived by annotating the answer span of the corresponding human-authored question for the paragraph, and isolating the sentence containing that span. We then provided each generated question and corresponding paragraph to the \textsc{BERT Shared-Norm MegaQA} model described in Section \ref{qa}. The results for each QG model in terms of answer F1 accuracy are shown in Table \ref{qa_on_qg}, compared alongside the result for human-authored questions.

As shown, the questions generated by the \textsc{RuleMimic} model are much better at eliciting the designated answers than the \textsc{Standard} model questions, indicating that observing the rule-generated questions alone is impactful. Additionally, the \textsc{Augmented} model generates more answerable questions than the \textsc{RuleMimic} model, showing the usefulness of combining rule-generated questions with human-authored questions as a data augmentation strategy.

\begin{table}[th!]
\begin{center}
\begin{tabular}{ | l | l | l | l | }
\hline
\textbf{Model} & \textbf{F1} \\
\hline
\textsc{Standard} & 0.354 \\
\textsc{RuleMimic} & 0.503 \\
\textsc{Augmented} & 0.551 \\
\textsc{Human} & 0.718 \\
\hline
\end{tabular}
\caption{Accuracy of QA system on QG output}
\label{qa_on_qg}
\end{center}
\end{table}

\subsubsection{Human Evaluation}

\begin{table}[th!]
\begin{center}
\begin{tabular}{ | m{0.275\linewidth} | m{0.15\linewidth} | m{0.15\linewidth} | }
\hline
\textbf{Model} & \textbf{Rating} & \textbf{Answer \vspace{0.075cm} Present} \\
\hline
\textsc{Standard} & 2.813 & 0.225 \\
\textsc{RuleMimic} & 2.934 & 0.381 \\
\textsc{Augmented} & 3.140 & 0.399 \\
\textsc{Human} & 3.776 & 0.793 \\
\hline
\end{tabular}
\caption{Human assessment of QG output}
\label{human_qg_eval}
\end{center}
\end{table}

We also elicited human judgments for a subset of the same generated questions. Participants were recruited from an internal team of linguists as well as Amazon Mechanical Turk (AMT). We selected questions corresponding to 175 inputs. Table \ref{qg_examples_table} in the appendix shows examples of these items. Participants read the input sentence in its paragraph context, then observed all four questions associated with the input (one generated by each of the three models plus the corresponding human-authored question). The presentation order of the questions for a given paragraph was randomized. Participants rated the fluency and answerability of questions on a scale of 1-4 based on the following statements: 

\textbf{1:} \textit{Question is completely ungrammatical. It's impossible to know what this question is asking.}

\textbf{2:} \textit{Question is mostly grammatical, but it doesn't fully make sense. It's not clear what this question is asking.}

\textbf{3:} \textit{Question is strangely worded, vague, or contains errors. However, I can make a guess about what the question is asking.}

\textbf{4:} \textit{Question is clearly worded. I understand what this question is asking.}

If the participant indicated that the question was answerable by rating it a 3 or 4, they were then asked if the answer to the question was contained in the paragraph. If they indicated `yes', they were asked to verify this by selecting all text spans in the paragraph that qualified as correct answers to the question. Based on this, we scored a question as having an `answer present' if it was marked as being answerable and if at least one of the participant-selected answer spans was the same one the question was conditioned upon when generated (signifying that the question actually elicited the answer the model observed in the input). 41 participants assessed a total of 1,560 paragraph-question items, with each item being rated by at least two participants (see Appendix \ref{qg_eval_details} for inter-rater reliability statistics). We averaged the scores for the same questions across participants. Table \ref{human_qg_eval} shows the mean ratings and answer presence for each set of generated questions including the \textsc{Human} questions. In terms of ratings, the results follow the same pattern as the automated evaluation: the \textsc{RuleMimic} questions are rated higher than the \textsc{Standard} questions, and the \textsc{Augmented} questions are rated higher than the \textsc{RuleMimic} questions. All sets of generated questions are rated much lower than the \textsc{Human} questions. The models are ordered the same in terms of answer presence, though the difference between the \textsc{RuleMimic} and \textsc{Augmented} models is slight. Overall these results again show the benefit of augmenting the training data with automatically generated questions. Accordingly, our demo currently runs the \textsc{Augmented} model.

\section{Generating Q\&A Pairs}\label{qna_section}

We combined our best-performing QG and QA models into a system that takes a text as input and returns a list of Q\&A pairs. Our web demo illustrates this system (see Appendix \ref{ui_details} for details). 

For our evaluations in Section \ref{qg}, the QG models observed annotated answer spans upon which the generated questions were conditioned. However, these annotations are obviously not available by default for any arbitrary text. Consequently, after splitting the text into sentences, we automatically identify syntactic chunks and named entities as candidate answers to questions (see Appendix \ref{qg_candidates_details} for details). For each candidate answer in a sentence, we produce an input consisting of that sentence annotated with that span. 
%(e.g. a sentence with 5 candidate answers yields 5 different annotated inputs)
We also include sentences with no answer annotations as inputs, since they are not formally required by the model. We provide all inputs for a given sentence to the \textsc{Augmented} QG model to get a list of questions that can be passed to the QA component. Note that even though some of the questions are already associated with annotated answers, we still apply QA as an additional means of verifying their answerability, and defer to the QA-predicted answer. To prepare the QA inputs, for each sentence-question item, we extend the sentence to include the sentences immediately preceding and following it, so each question becomes aligned with a 3-sentence passage. This enables the QA system to possibly retrieve additional context beyond the sentence that it may deem as part of the answer span.
%(particularly useful for the QG inputs without answer annotations)
We provide these passage-question pairs to the \textsc{BERT Shared-Norm MegaQA} model, then retain output items for which answers are found. We reduce the redundancy of items by filtering those with duplicate questions or answers, as well as items where the question and answer concatenated share 60\% or more of the same tokens. In these cases, we only retain the item with the highest QG probability. 

% \deleted{To reduce redundancy, we also retain only the first item among items that have the same question, the same answer, or pertain to the same original input sentence. We display the resulting Q\&A pairs to the user.}

\subsection{Human Evaluation}\label{qna_eval_section}

\begin{figure}[h!]
\centering
\includegraphics[width=\linewidth]{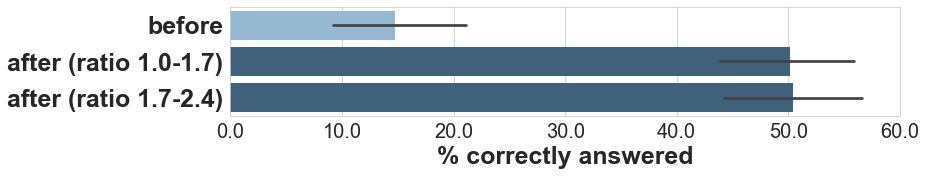}\\
\caption{Human accuracy on target questions before and after observing generated Q\&A pairs}
%, with the 'after' result grouped by the number of generated items participants observed relative to the number of target questions
\label{qna_pairs_eval}
\end{figure}

We used our system to generate Q\&A pairs for ten texts from the \textsc{SQuAD} test set. Appendix Table \ref{qna_examples_table} shows an example of a generated Q\&A list for one text. We conducted an evaluation of the informativeness of these pairs with 38 AMT participants. In the first stage of the evaluation, participants were shown only the title of one text (e.g. ``Tesla'') and the human-authored \textsc{SQuAD} questions (no answers) corresponding to the text. Without referencing any material, they were asked to answer these target questions or respond with ``X'' for questions they couldn't answer. Because no generated Q\&A pairs were shown to participants during this stage, the accuracy of their answers indicated their prior mental knowledge of the information in the text. In the second stage, the generated Q\&A pairs for the same text were revealed to them and they answered the same target questions again. Participants never observed the original text itself. The logic of this design is that the more questions people could correctly answer in the second stage relative to the first, the more informative the generated Q\&A list could be deemed. The ratio of generated Q\&A pairs to target questions per text varied from 1 to 2.4 (e.g. ratio = 2 for a text with 30 generated pairs and 15 target questions). Figure \ref{qna_pairs_eval} shows the percentage of target questions participants answered correctly before and after observing the Q\&A list, grouped by ratio. The overall difference in these conditions (14.74\% vs. 50.26\%) shows that the generated items were partially informative for answering the target questions, signifying that the system does highlight some of the same content people ask questions about. However, accuracy did not markedly improve as participants saw more items (50.18\% for the lower ratio vs. 50.38\% for the higher ratio), suggesting that the information coverage of the items could be improved. See Appendix \ref{qna_evaluation_details} for more details about this evaluation.

% \deleted{\section{Demo Interaction}\label{demo_interaction}

% The system is a web application (see Appendix \ref{ui_details}) where the user can provide any text as input in the interface, then automatically generate Q\&A pairs for the text. The application additionally includes a component where users can author and obtain answers for their own custom questions (see Appendix \ref{interactive_qa}).}

\section{Conclusion and Future Work}\label{conclusion}

In this paper, we present a system that automatically produces Q\&A pairs for multi-paragraph documents. We report some novel experiments for QA and QG that motivate techniques for improving these tasks. We show that combining these components can produce informative Q\&A items. Our future work will focus on more advanced modeling of information structure in documents. For example, the ideal design of Q\&A items varies by domain (e.g. news stories vs. financial reports vs. opinion editorials), and items should target the content readers find most substantial in each domain.

% \section*{Acknowledgments}

% The acknowledgments should go immediately before the references. Do not number the acknowledgments section.
% Do not include this section when submitting your paper for review.

\bibliography{qna_demo}
\bibliographystyle{acl_natbib}

\newpage

\appendix

\section{Appendices}\label{appendix}

\subsection{QA Model Details}\label{qa_details}

The TF-IDF method for ranking paragraph relevance to the question specifically uses the BM-25 ranker\footnote{\href{https://pypi.org/project/rank-bm25/}{pypi.org/project/rank-bm25}} \cite{robertson2009probabilistic}. We implemented the QA model in PyTorch using the HuggingFace Transformers library\footnote{\href{https://huggingface.co/transformers/index.html}{huggingface.co/transformers}}. As described in Section \ref{qa_overview}, we use the pre-trained \textsc{BERT-Base-Uncased} model, which has 12 layers, 768 nodes per layer, 12 heads per layer, and 110M parameters overall. The maximum sequence length for \textsc{BERT-Base-Uncased} is 384 tokens (including both paragraph and question tokens combined), so we truncated paragraphs when this total was exceeded. The output layer consists of a 384 x 2 matrix whose dimensions correspond to token indices for the start and end of the answer span. We trained the model in parallel across 4 Nvidia Tesla V100 GPUs with a paragraph-question batch size of 48 with gradient accumulation at step 1 (12 paragraph-question pairs per GPU, which was the maximum size a single V100 GPU could accommodate). Following \citeauthor{devlin-etal-2019-bert}'s \textsc{BERT}-based fine-tuning procedure for paragraph-level QA, the model was trained for 3 epochs and a learning rate of 3e-5 using Adam optimization.

\subsection{QG Model Details}\label{qg_details}

We used OpenNMT-py\footnote{\href{https://github.com/OpenNMT/OpenNMT-py}{github.com/OpenNMT/OpenNMT-py}}  \cite{opennmt} for implementation of the QG model. For BPE tokenization, we use the OpenAI GPT-2 tokenizer implemented by the HuggingFace transformers library cited above. The vocabulary included all tokens observed in the training data. The Transformer encoder and decoder each consist of 4 layers with 2048 nodes and 8 heads each. We include position encodings on the token embeddings and a copy attention layer in the decoder. We used a training batch size of 4096 tokens, normalizing gradients over tokens and computing gradients based on 4 batches. We trained for a maximum of 100,000 steps and validated every 200 steps, with early stopping after one round of no improvement in validation loss. We applied the other hyperparameter settings recommended for training transformer sequence-to-sequence models on the OpenNMT-py website\footnote{\href{https://opennmt.net/OpenNMT-py/FAQ.html\#how-do-i-use-the-transformer-model}{opennmt.net/OpenNMT-py/FAQ.html\#how-do-i-use-the-transformer-model}}. This included Adam optimization with $\beta_1$ = 0.998, gradient re-normalization for norms exceeding 0, Glorot uniform parameter initialization, 0.1 dropout probability, noam decay, 8000 warmup steps for decay, learning rate = 2, and label smoothing $\epsilon$ = 0. 

% \subsection{QG Output}\label{qg_examples_section}

% Table \ref{qg_examples_table} shows several examples of the sentence-question pairs generated by each model.

\subsection{QG Evaluation Details}\label{qg_eval_details}

The sentence inputs for the evaluated questions were randomly sampled after filtering for those inside paragraphs longer than 500 characters, to ensure participants could efficiently complete the evaluation. AMT workers were paid \$7 for their participation in this evaluation, with the expected time commitment of about 35 minutes.

The Cohen's kappa inter-rater agreement on the fluency/answerability ratings of 1-4 was 0.422, indicating moderate agreement. The kappa for answer presence in the paragraph was 0.465, also indicating moderate agreement. 

\subsection{System Implementation Details}\label{ui_details}

\begin{figure*}[h!]
\begin{minipage}{\textwidth}
\centering
\includegraphics[width=\linewidth]{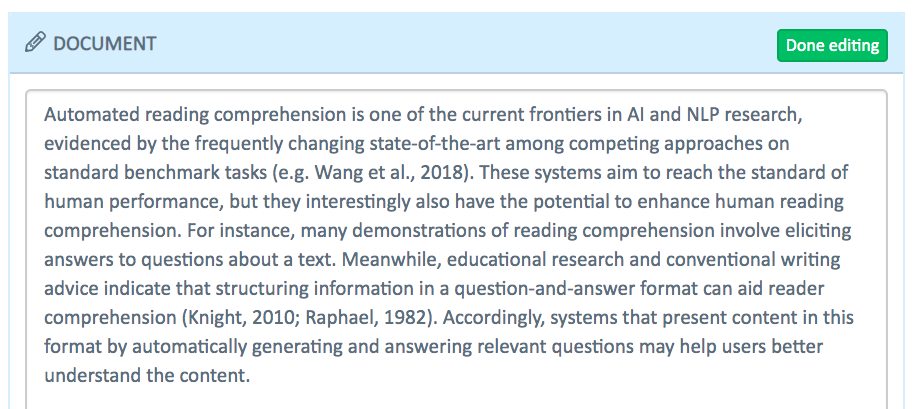}\\
\includegraphics[width=\linewidth]{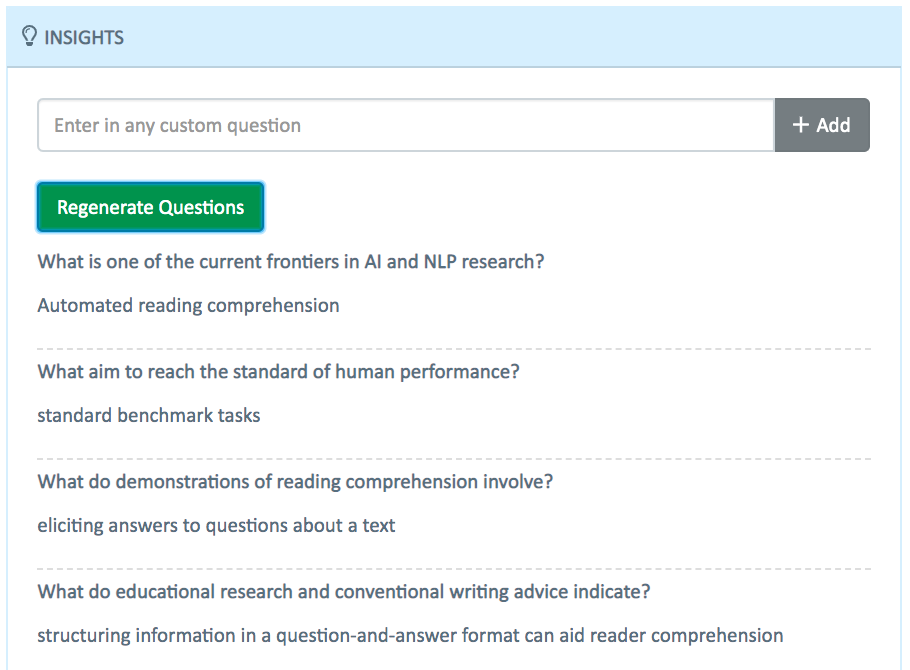}
\end{minipage}
\caption{Screenshot of UI with generated Q\&A pairs for the displayed document}
\label{screenshot}
\end{figure*}

The system UI is implemented using React JS with Bootstrap CSS for styling. Figure \ref{screenshot} shows a screenshot of the interface. The QA and QG functionalities run as web services implemented using Flask.

As an additional feature of the UI, users have the option to obtain answers to their own custom questions. They supply the question via a text box. The QA system receives the entire document text as input along with the question. We enforce paragraph boundaries by splitting the document into non-overlapping paragraphs of 300 tokens, and then apply the \textsc{BERT Shared-Norm MegaQA} model with top k=4 for paragraph retrieval\footnote{Section \ref{qa_comparison_section} reported the result for k=15, but k=4 performs only slightly lower (71.21 EM and 78.89 F1 vs. 72.85 and 80.58, respectively) with significantly higher efficiency.}. If the model predicts the answer is not in the text, the user sees a message indicating this.

\subsection{Candidate Answers for QG}\label{qg_candidates_details}

We use the spaCy\footnote{\href{https://spacy.io/}{spacy.io}} library to extract all named entities and noun chunks. Additionally, we extract all dependency parse subtrees whose head is labeled as one of the following: clausal complement (xcomp), attribute (attr), prepositional modified (prep), object (obj), indirect object (iobj), flat multiword expression (flat), fixed multiword expression (fixed), clausal subject (csubj), clausal complement (ccomp), adjectival clause (acl), and conjunct (conj). All extracted chunks are annotated as answer spans.

\subsection{Q\&A List Evaluation}\label{qna_evaluation_details}

We truncated each of the ten \textsc{SQuAD} documents to its first three paragraphs. There were on average 334.5 tokens per truncated document. For each document we selected all \textsc{SQuAD} questions corresponding to the first three paragraphs as the list of target questions participants were prompted to answer. There were on average 16.2 target questions per truncated document. We provided the truncated document to the system to generate a list of Q\&A items.  As described in Section \ref{qna_eval_section}, the ratio of generated Q\&A items per target questions varied from 1 to 2.4 with an average of 1.66, resulting in an overall average of 26.3 generated items per document. 

Each of the 38 AMT participants answered the target questions for a single document, so approximately four participants answered each unique list of target questions. They were paid \$8 for their participation, with the expected time commitment of around 30 minutes. The instructions emphasized that they should not use any external information to answer the questions other than the reference Q\&A list (which was only used when participants answered the questions for the second time). They were told their participation would not be judged based on the number of questions correctly answered. Participants were not informed that the reference Q\&A items were automatically generated.

Because all answers were provided as free text and there could be some token variation in correct answers for the same question (e.g. ``Parliament of the United Kingdom'' vs. ``UK Parliament''), we used a fuzzy metric for judging answers as correct. We counted a participant answer as correct if it had at least one token in common with the answer given in the \textsc{SQuAD} dataset. This is a permissive threshold that can yield false positives (e.g. ``300 years'' vs. ``500 years''), but because it was consistently applied across both stages of the evaluation (i.e. before and after observing the Q\&A list), we deemed it sufficient for quantifying the relative impact of the generated items in the `after' condition.

\onecolumn

\begin{table*}[th!]
\begin{center}
\begin{tabular}{ | p{0.325\linewidth} | l | p{0.45\linewidth} | }
\hline
\textbf{Input Sentence} & \textbf{Model} & \textbf{Output Question} \\
\hline
\multirow{4}{0.325\textwidth}{By far the most famous work of Norman art is the \texttt{$<$ANSWER$>$} Bayeux Tapestry \texttt{$<$/ANSWER$>$}, which is not a tapestry but a work of embroidery.} & \textsc{Standard} & What is the name of the work of Norman art?	\\
% \cline{2-3}
& \textsc{RuleMimic} & What is not a tapestry but a work of embroidery? \\
& \textsc{Augmented} & What is the most famous work of Norman art?\\
& \textsc{Human} & What is Norman art's most well known piece?\\ 
\hline
\multirow{4}{0.325\textwidth}{Later in life, Tesla made claims concerning a "teleforce" weapon after studying the \texttt{$<$ANSWER$>$} Van de Graaff generator \texttt{$<$/ANSWER$>$}.} & \textsc{Standard} & What was the name of the weapon that studying the Teslaforce?\\
& \textsc{RuleMimic} & What did Tesla make claims concerning a "teleforce" weapon after studying?\\
& \textsc{Augmented} & What did Tesla study?\\
& \textsc{Human} & What was he studying that gave him the teleforce weapon idea?\\
\hline
\multirow{4}{0.325\textwidth}{The Super Bowl 50 Host Committee has vowed to be ``the most giving Super Bowl ever'', and will dedicate \texttt{$<$ANSWER$>$} 25 percent \texttt{$<$/ANSWER$>$} of all money it raises for philanthropic causes in the Bay Area.} & \textsc{Standard} & How much of the Super Bowl raises?\\
& \textsc{RuleMimic} & What will the Super Bowl 50 Host Committee dedicate it raises for philanthropic causes in the Bay Area?\\
& \textsc{Augmented} & What is the Super Bowl 50 Host Committee?\\
& \textsc{Human} & How many will the host committee dedicate to local charities?\\
\hline
% \multirow{4}{0.325\textwidth}{On \texttt{$<$ANSWER$>$} 13 June 1525 \texttt{$<$/ANSWER$>$}, the couple was engaged with Johannes Bugenhagen, Justus Jonas, Johannes Apel, Philipp Melanchthon and Lucas Cranach the Elder and his wife as witnesses.} & \textsc{Standard} & When did Johannes Apel die?\\
% & \textsc{RuleMimic} & When was the couple engaged with Johannes Bugenhagen on 13?\\
% & \textsc{Augmented} & When was the couple engaged?\\
% & \textsc{Human} & When were Luther and his prospective bride engaged? \vspace{27}\\
% \hline
\multirow{4}{0.325\textwidth}{In 1899, John Jacob Astor IV invested \$100,000 for Tesla to further \texttt{$<$ANSWER$>$} develop and produce a new lighting system \texttt{$<$/ANSWER$>$}.} & \textsc{Standard} & What did Jacob Astor IV do?\\
& \textsc{RuleMimic} & What did John Jacob Astor IV invest \$100,000 for in 1899?\\
& \textsc{Augmented} & Why did Jacob Astor IV invest \$100,000?\\
& \textsc{Human} & What did Astor expect the money be used for?\\
\hline
\multirow{4}{0.325\textwidth}{Most influential among these was the definition of Turing machines by Alan Turing in \texttt{$<$ANSWER$>$} 1936 \texttt{$<$/ANSWER$>$}, which turned out to be a very robust and flexible simplification of a computer.} & \textsc{Standard} & When was the definition of the definition of Turing?\\
& \textsc{RuleMimic} & When turned out to be a very robust and flexible simplification of a computer? \\
& \textsc{Augmented} & When did Alan Turing write machines?\\
& \textsc{Human} & In what year was the Alan Turing's definitional model of a computing device received?\\
\hline
\multirow{4}{0.325\textwidth}{In addition to the \texttt{$<$ANSWER$>$} Vince Lombardi \texttt{$<$/ANSWER$>$} Trophy that all Super Bowl champions receive, the winner of Super Bowl 50 will also receive a large, 18-karat gold-plated ``50''.} & \textsc{Standard} & What is the name of the Super Bowl?\\
& \textsc{RuleMimic} & Who will the winner of Super Bowl 50 also receive a large in addition to the Vince Lombardi Trophy that all Super Bowl champions receive? \\
& \textsc{Augmented} & Who wrote the Super Bowl 50?\\
& \textsc{Human} & Who is the trophy given to the Super Bowl champion named for?\\
\hline
\multirow{4}{0.325\textwidth}{In 1874, Tesla evaded being drafted into the Austro-Hungarian Army in Smiljan by running away to \texttt{$<$ANSWER$>$} Tomingaj \texttt{$<$/ANSWER$>$}, near Gra\v{c}ac.} & \textsc{Standard} & What was the name of Tesla's Army in 1874?\\
& \textsc{RuleMimic} & Who was near Gra\v{c}ac? \\
& \textsc{Augmented} & Where did Tesla travel to?\\
& \textsc{Human} & Where did Tesla run to avoid the army draft? \vspace{15pt}\\
\hline
\end{tabular}
\caption{Examples of questions produced by each evaluated QG model for the given input sentences}
\label{qg_examples_table}
\end{center}
\end{table*}

\begin{table*}[th!]
\begin{center}
\begin{tabular}{|p{0.45\linewidth}|p{0.45\linewidth}|} %| p{0.45\linewidth}
%\textbf{Input Sentence} & \textbf{Model} & \textbf{Output Question} \\
\hline
Q: What is separate from the combustion products? \newline
A: working fluid &
Q: Where was the water supply for driving waterels?\newline
A: factories\\
\hline
Q: What is solar power?\newline
A: Non-combustion heat sources &
Q: What did the mine provide?\newline
A: water supply\\
\hline
Q: What is the ideal thermodynamic cycle used for?\newline
A: to analyze this process &
Q: Where was it employed?\newline
A: draining mine workings\\
\hline
Q: What is heated and transforms into steam?\newline
A: water&
Q: Where was the storage reservoir?\newline
A: above the wheel\\
\hline
Q: Why is mechanical work done?\newline
A: When expanded through pistons or turbines&
Q: What was passed over the wheel?\newline
A: Water\\
\hline
Q: What is then condensed and pumped back into the boiler?\newline
A: reduced-pressure steam&
Q: When was the first railway journey?\newline
A: 21 February 1804\\
\hline
Q: Who invented the first commercially true engine?\newline
A: Thomas Newcomen&
Q: Where was the train?\newline
A: along the tramway from the Pen-y-darren ironworks, near Merthyr Tydfil to Abercynon in south Wales\\
\hline
Q: What could generate power?\newline
A: atmospheric engine&
Q: What was built by Richard Trevithick?\newline
A: The first full-scale working railway steam locomotive\\
\hline
Q: Who proposed the piston pump?\newline
A: Papin&
Q: The design incorporated a number of what?\newline
A: important innovations that included using high-pressure steam which reduced the weight of the engine and increased its efficiency\\
\hline
Q: What happened to Newcomen's engine?\newline
A: relatively inefficient&
Q: What did England become the leading centre for?\newline
A: experimentation and development of steam locomotives\\
\hline
Q: What was the engine used for?\newline
A: pumping water&
Q: Where was the railways colliery?\newline
A: north-east England\\
\hline
Q: What was the vacuum worked by?\newline
A: condensing steam under a piston within a cylinder&
Q: Who visited the Newcastle area in 1804?\newline
A: Trevithick\\
\hline
Q: What was the reason for draining waterelswheels?\newline
A: providing a reusable water supply &\\
\hline
\end{tabular}
\caption{Generated Q\&A list for the first three paragraphs of the \textsc{SQuAD} document titled ``Steam engine''}
\label{qna_examples_table}
\end{center}
\end{table*}

\end{document}